\documentclass[letterpaper, 10 pt, conference]{ieeeconf}  % Comment this line out if you need a4paper

\IEEEoverridecommandlockouts                              % This command is only needed if 
\usepackage{times}
\usepackage[OT1]{fontenc}

\usepackage{multicol}
\usepackage{url}
\usepackage[pdftex]{graphicx}
\usepackage{subcaption}
\usepackage{algorithm}
\usepackage{amsmath}
\usepackage{amssymb}
\usepackage{amsfonts}
\usepackage{algcompatible}
\usepackage{framed}
\usepackage{balance}
\usepackage{bm}
\usepackage{ulem}
\usepackage{dsfont}
\usepackage{comment}
\usepackage{multirow}
\usepackage{hyperref}

\newcommand{\Tau}{\mathcal{T}}

\usepackage{color}
\usepackage{xcolor}

\title{\LARGE \bf
One-Shot Learning of Multi-Step Tasks from Observation via \\ Activity Localization in Auxiliary Video
}

\author{
    Wonjoon Goo and Scott Niekum\\
    Department of Computer Science\\
    University of Texas at Austin, Austin, TX 78712\\
    \texttt{\{wonjoon, sniekum\}@cs.utexas.edu}\\
}

\begin{document}

\setlength{\skip\footins}{0.3cm}

\maketitle
\thispagestyle{empty}
\pagestyle{empty}

%%%%%%%%%%%%%%%%%%%%%%%%%%%%%%%%%%%%%%%%%%%%%%%%%%%%%%%%%%%%%%%%%%%%%%%%%%%%%%%%
\begin{abstract}
Due to burdensome data requirements, learning from demonstration often falls short of its promise to allow users to quickly and naturally program robots. Demonstrations are inherently ambiguous and incomplete, making correct generalization to unseen situations difficult without a large number of demonstrations in varying conditions.
By contrast, humans are often able to learn complex tasks from a single demonstration (typically observations without action labels) by leveraging context learned over a lifetime. Inspired by this capability, our goal is to enable robots to perform one-shot learning of multi-step tasks from observation by leveraging auxiliary video data as context.  
Our primary contribution is a novel system that achieves this goal by: (1) using a single user-segmented demonstration to define the primitive actions that comprise a task, (2) localizing additional examples of these actions in unsegmented auxiliary videos via a metalearning-based approach, (3) using these additional examples to learn a reward function for each action, and (4) performing reinforcement learning on top of the inferred reward functions to learn action policies that can be combined to accomplish the task. 
We empirically demonstrate that a robot can learn multi-step tasks more effectively when provided auxiliary video, and that performance greatly improves when localizing individual actions, compared to learning from unsegmented videos.

% \SN{NOTE: Maybe we should have another experiment that does the IRL+RL step, but only with the data from the original demo. It will be very bad, but it shows the value of the auxiliary videos.  Also, I think we should change how we talk about shuffle and learn.  It isn't really IRL since there are no actions -- it is just reward function inference.}

\end{abstract}

%%%%%%%%%%%%%%%%%%%%%%%%%%%%%%%%%%%%%%%%%%%%%%%%%%%%%%%%%%%%%%%%%%%%%%%%%%%%%%%%
\section{Introduction}

Learning from demonstration (LfD)~\cite{argall2009survey} has emerged as a powerful way to quickly and naturally program robots to perform a wide variety of tasks.  Unfortunately, demonstrations are inherently ambiguous and incomplete. Correct generalization to unseen situations is therefore difficult without a large number of demonstrations in varying conditions. This data requirement places a significant burden on end-users, often limiting the use of LfD to simple tasks.
 
By contrast, humans are often able to learn complex tasks from a single demonstration by leveraging context learned over a lifetime---for example, information about how objects work, episodic memories of similar situations, or an intuitive understanding of the intentions of the demonstrator. Similarly, robots increasingly have access to auxiliary sources of video data---for example, from prior experiences, curated datasets such as the Epic-Kitchen dataset \cite{Damen2018EPICKITCHENS}, or less structured Youtube videos. In this work, we propose to leverage auxiliary video data as contextual information to help robots intelligently disambiguate and generalize a single demonstration of a multi-step task. Essentially, a single user-provided, segmented demonstration describes \textit{what} activities to perform (as well as one example of how to perform them, grounded in the actual environment that the robot will act in), while the auxiliary video data provides additional examples of \textit{how} each activity ought to be performed, allow the robot to learn to generalize without overfitting.

% Prior work has looked at using video in various ways for LfD/LfO, but has not addressed complex, multi-step tasks sufficiently
Although prior works have explored the use of video data in an LfD setting, in all instances that we are aware of these methods have only addressed single step tasks \cite{unsup-reward-for-imitation-sermanet,TCN2017-sermanet} or have assumed well-aligned data with little variance \cite{imitation-from-obsv-liu}.  However, many common robotics tasks, such as cooking and assembly, require multiple steps that may have different goals and involve different objects or features.  
Thus, as we show experimentally (and as other works have argued
\cite{niekum2012learning,konidaris2012robot}), learning a separate policy for each step of a task can lead to improved generalization.

Our primary contribution is a novel framework that can localize additional examples of each user-demonstrated action in unsegmented auxiliary videos, which are then used to aid learning.
Specifically, we cast the problem of action localization as a single-shot activity recognition problem, in which we only have one example of each activity (from user-provided demonstration segments) and attempt to classify small sets of frames in each of the auxiliary videos as one (or none) of those activities.

\begin{figure*}[t]
    \centering
    \includegraphics[width=0.85\linewidth]{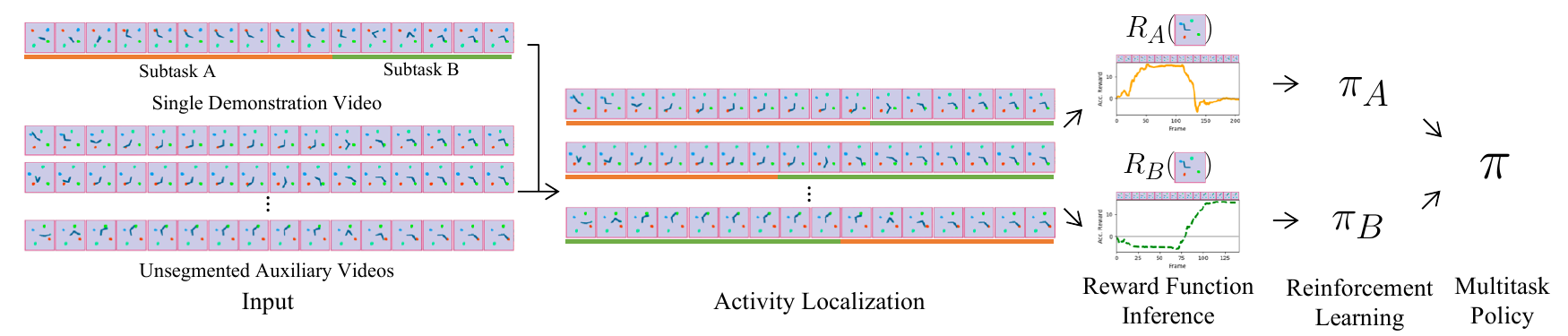}
    \caption{
        The proposed one-shot learning from observation pipeline: (1) A single segmented demonstration defines subtasks; (2) additional clips of each subtask are then localized in unsegmented auxiliary videos; (3) these clips are used to infer a reward function for each subtask; (4) RL is used to learn subtask policies that can be combined to complete the full task.
    }
    \label{fig:whole_pipeline}
    \vspace{-0.5cm}
\end{figure*}

% Introduce LfO and compare/contrast to LfD
We then use the segmented video clips in tandem with the original demonstration segments to learn to perform each step of the task separately. However, one significant difficulty in utilizing video data in an LfD setting is that there are typically no available action labels for the observations. This difficulty has motivated work in the learning from observation (LfO) setting. Observations without action labels are generally not sufficient for direct imitation learning, but have been used instead to help build dense reward signals \cite{imitation-from-obsv-liu, unsup-reward-for-imitation-sermanet}, learn object affordances \cite{lopes2007affordance}, or resolve ambiguities of written instructions \cite{misra2014tell}.  

While there are many possible ways to use the segmented video data from our algorithms in an LfO setting, we focus on an inverse reinforcement learning setting.  Specifically, for each subtask, the segmented video clips of that activity are used to perform reward function inference, followed by reinforcement learning 
%(via Proximal Policy Optimization \cite{ppo-schulman2017}) % Too much detail for intro
on the inferred reward functions.  The learned policies can then be sequentially executed to accomplish the task in novel situations.  Figure \ref{fig:whole_pipeline} illustrates this full learning pipeline.

We first demonstrate the accuracy of our one-shot activity segmentation algorithm on both well-constrained, simulation-generated videos and unconstrained, naturalistic videos, namely the ActivityNet \cite{caba2015activitynet} dataset. 
%Generalization capabilities are also shown via intra-domain and out-of-domain tests.  
%Additionally, in a simulated multi-step two-joint reaching task, we show that our segmentation-based system significantly outperforms reinforcement learning on reward functions inferred from unsegmented video, as well as learning from a single demonstration without auxiliary video. 
Additionally, we perform reinforcement learning experiments in a simulated multi-step two-joint reaching task, in which the reward function is inferred from demonstration video data. We empirically demonstrate that the robot can learn more effectively when provided auxiliary video data, and that performance greatly improves when localizing individual actions, compared to learning from unsegmented videos.

%\WJ{
%(Notes: I want to emphasize once again, reminding the readers that our system as a whole is our main contribution.)
%This whole pipeline performing one-shot multi-step learning from observation algorithm is the main contribution.
%}  % SCOTT: I don't think it is necessary, or even good, to say it again
%We anticipate that this work will be a step toward more comprehensive LfO algorithms that will eventually leverage unstructured, web-scale video data (and perhaps other modalities thereof) to enable robust one-shot (or few-shot) learning from demonstration.

%%%%%%%%%% OUTLINE FOR REST OF PAPER %%%%%%%%%%%%%%%

% 1) Problem description

\section{Related Work}
Learning from observation (LfO) is a recent area of research that aims to perform imitation learning on general sensory inputs, such as visual signals, without access to the true state or action labels \cite{imitation-from-obsv-liu,unsup-reward-for-imitation-sermanet,TCN2017-sermanet}. Imitation learning has been performed in an LfO setting by training a context translation network by using a feature tracking loss in image space \cite{imitation-from-obsv-liu}, by finding the most discriminative features directly related to subgoals \cite{unsup-reward-for-imitation-sermanet}, or by inferrring a reward function via a self-supervised embedding network \cite{TCN2017-sermanet}. However, all of these approaches assume pre-aligned videos of single-step tasks, restricting their applications. 
%\SN{\sout{I didn't totally understand some of the language you used here, so make sure that I didn't make anything incorrect with my edits.}}

Recently, several one-shot imitation learning methods have been proposed to obtain high sample efficiency in the LfO setting \cite{one-shot-visual-finn2017,one-shot-imit-duan}. While other LfO approaches typically require performing reinforcement learning on an inferred reward function, one-shot imitation learning generates a policy directly after observing a single demonstration. Recent examples have utilized attention-based models \cite{one-shot-imit-duan} and meta-learning methods \cite{one-shot-visual-finn2017}.
%This is achieved by specialized neural network architectures, such as an attention-based model [cite] or a meta-learning method [cite]. 
%Still, these approaches can only generalize across different instance of a task, not across fundamentally different tasks. The suggested action localization problem and following LfO pipeline aims to learn an unseen multi-step task while pursuing the same goal of sample efficiency.
However, these methods are restricted to single-step tasks within a distribution of task instances similar to those seen in the training data.

The action localization problem has been well-studied in the computer vision literature \cite{poppe2010survey,karpathy2014large}. The most closely related work is that of \cite{Yang_2018_CVPR}, in which they propose a similar few-shot action localization problem and solve it through a meta-learning framework. 
%\WJ{Since their method is built on top of the specific meta-learning framework, Matching Network \cite{vinyals2016matching}, it requires a specialized neural network called full context embedding network while our approach does not.} 
However, their approach requires a specialized network architecture---a full context embedding network---whereas our approach is fully general, allowing the flexibility of choosing any network architecture.
Additionally, though we share the same goal of one-shot action localization, our ultimate goal is to apply learning from observation on top of action-localized multi-step videos.

The work of Hausman et al. \cite{mm-imit-learn-karol2017} is also closely related to our problem setting. Similar to our goal of segmenting videos involving multiple subtasks, they suggest a method that can discover skills by inferring latent categorical codes through an InfoGAN-like \cite{infogan-chen2016} formulation. Although their method successfully imitates multiple policies for different skills with a single neural network, it requires state-action pairs, rather than purely observational data. By contrast, our LfO approach can be applied to videos without any action labels.

\section{One-Shot Multi-Step Task Learning}
We address the one-shot multi-step task learning problem in two parts, which we describe in the following subsections: (1) a one-shot activity localization algorithm, followed by (2) an LfO algorithm that is used separately for each subtask. 

%The activity localization algorithm is used for bootstrapping the number of videos for each specific step by mining videos from unparsed auxiliary dataset. After we run a LfO algorithm with bootstrapped demonstration videos for each step to get a policy solving each subtask. The difficulty of learning from raw observation can be greatly reduced by, first, having well aligned demonstration videos, and second, focusing on specific subtask.

\subsection{One-Shot Activity Localization}

%One-shot activity localization solves the problem of LfO algorithms, which require a large amount of well-preprocessed data; by searching over large unparsed videos and finding only relevant part to a given sub-step task demonstration, the quantity and quality data requirement can be fulfilled effortlessly. Thereby, we achieve the sample efficiency even we face unseen new task.

LfO algorithms typically require large amounts of well-aligned, preprocessed examples of a task (or steps of a task, in our case).  Rather than relying on this, we assume that we have only a single, user-segmented demonstration of the task.  We then propose to use a one-shot activity localization algorithm that can identify clips of activities in auxiliary videos that match the activities in each of the segments of the demonstration, providing additional examples of each.  Thus, given a source of unsegmented auxiliary task-relevant videos, this approach can automatically gather activity-level video data for any new task.

% Mathematical problem formulation
%Formally, we define a one-shot activity localization problem as follows. Assume that a demonstration video $\mathbf{v}^d = [v_1, \dots ,v_n]$ is given for a K-step task $\tau$ with following step-label $\mathbf{l}^d = [l_1, \dots ,l_n] \in [1, \dots ,K]^n$. Then, given a query video $\mathbf{v}^q$, we want to generate a dense label of steps so that we can group snippets belong to a same step and use group of snippets for LfO framework separately. 
In this paper, a video $\mathbf{v}$ is treated as a sequence of short snippets (typically of some small fixed length, e.g. 5 frames): $\mathbf{v} = [v_1, v_2, ..., v_t]$. A demonstration video is given in which $K$ different activities are shown; the demonstration snippets have associated labels $\mathbf{l}^{demo} = [l_1, \dots ,l_t] \in [1, \dots ,K]^t$ corresponding to the activity that each one belongs to. For some new unlabeled target video $\mathbf{v}^{target}$, action localization can be defined as a classification problem in which a classifier assigns a label to each of the snippets in $\mathbf{v}^{target}$.  Thus, unlike a standard classification problem with a fixed set of labels, the number of labels (and the corresponding activities) are determined by the demonstration.  To address this classification problem in a one-shot manner, we use a framework based on model agnostic meta learning (MAML) \cite{maml-finn17a} and its first order approximated version, Reptile \cite{nichol2018reptile}, with a deep neural network.

% We want to solve this problem through deep learning, but there is a limitation. So, we will use meta learning framework.
%We resort to deep learning method in order to handle high-dimensionality of input. However, it is unrealistic to expect a neural network to be trained with only few given snippets for each step. Therefore, we used a meta-learning framework, namely model agnostic meta learning (MAML) \cite{maml-finn17a}.

MAML and Reptile aim to find good initial network parameters $\theta$ that can efficiently adapt to new problems with a single (or small number of) stochastic gradient descent steps. Parameters found with these frameworks are not for any specific problem; instead, MAML and Reptile directly optimize for performance of the network when the parameters are fine-tuned via SGD step(s) on a small amount of training data for a particular problem.  In our case, we wish to learn initial parameters that allow us to successfully train an activity classifier with only a single demonstration (a small number of snippets) of each activity.

Finding these initial parameters can be done through SGD as in other deep learning methods. MAML requires a set of training tasks $\tau \in \Tau$, and criteria $C_\tau$ to calculate the goodness of the parameters---how well the initial parameters perform after one SGD step on a new problem $\tau$. The gradient descent step is formally defined as:
\begin{equation}
    \theta \leftarrow \theta - \beta \nabla_\theta \sum_{\tau \in \Tau} C_{\tau}(\; \cdot \;;\theta_{\tau}),
    \label{eq:maml}
\end{equation}
where $\theta_{\tau} = \theta - \alpha \nabla_\theta C_{\tau} ( \cdot ;\theta)$, and $\alpha$ and $\beta$ are learning rates for finetuning and MAML training respectively.
%It is noteworthy that the generalization power of MAML comes from training for a \textit{set} of tasks, instead of a single task instance.

%Few-shot learning is a good example in which meta-learning framework can be applied. If we have multiple few-shot learning tasks, then we can find a good initial parameter through MAML framework. Then, at the testing time, a specialized parameter can be calculated with a single SGD step with small number of target task example. 

%Since we cast our problem as a few-shot learning problem, we can directly apply MAML framework for our video segmentation problem. 
In our setting, a classification problem $\tau$ is defined by the set of activity labels for a given demonstration video.
%if two videos sharing same activity labels $L_\tau$, then two videos are regarded as a same task. 
The loss function $C_\tau$ is a softmax cross entropy loss since the problem is a K-way classification problem.
Therefore, Eq.~\ref{eq:maml} is formally represented as:
\begin{equation}
    \theta \leftarrow \theta - \beta \nabla_\theta \sum_{ \tau \in \Tau } C_{\tau}(\bm{v}^{target}_\tau,\bm{l}^{target};\theta_{\tau}),
    \label{eq:maml-specific}
\end{equation}
where $\theta_{\tau} = \theta - \alpha \nabla_\theta C_{\tau} (\bm{v}^{demo}_{\tau},\bm{l}^{demo}_{\tau};\theta)$, and both $(\bm{v}^{demo}_{\tau},\bm{l}^{demo}_{\tau})$ and $(\bm{v}_{\tau}^{target},\bm{l}_{\tau}^{target})$ are data samples belonging to the same task $\tau$. 
%\SN{\sout{What are v and l here?  Are they whole segments? Snippets?  Whole videos?  Be more specific and unify this notation with the notation at the beginning of this sections}}

After we learn good initial parameters $\theta$ through Eq.~\ref{eq:maml-specific}, one-shot activity localization with a demonstration video can be performed using $\theta_{\tau}$, which are parameters that are fine-tuned with a demonstration video $\bm{v}^{demo}_{\tau}$ and labels $\bm{l}^{demo}_{\tau}$. A neural network of parameters $\theta_{\tau}$ is specialized for a specific task $\tau$, and it will emit a dense classification result for every snippet from a given task video $\bm{v}_{\tau}$. For the case in which a target video contains snippets of other activities not included in the task $\tau$, we set a matching threshold, such that if predictions for all the classes are below this value, we will instead output ``none of the above''.

% TODO: Later, connect back to LfO: You might be able to talk about how your work extends this to make it clear what your novel contribution is.

\subsection{Learning from Observation}
\label{sec:lfo}

% Breifly points out what will happen with pretrained neural network.
Now, for any given demonstration video, an activity localization network can be quickly trained in a one-shot fashion. 
%Specifically, we will extract subparts from a video in auxiliary video dataset for each different step. Then, a policy for each step will be trained separately.
Given an auxiliary set of task-relevant videos, this network can then be used to label each snippet of these videos as one (or none) of the demonstrated activities.  This provides additional examples of each activity, which will support learning a separate reward function and  policy for each activity. Since labeling happens at the snippet level, this process will not necessarily yield clean, contiguous activity segments. 
Future work may explore the advantages of de-noising these labels and identifying large, contiguous action segments, but we demonstrate that our approach to reward inference works even in the presence of moderate label noise.

% We need a special consideration since our activity localization is noisy.
%Since our problem formulation on activity localization is K-way classification (where K is the number of steps given in a demonstration video), the output prediction is made for each video snippet, and it does not provide segmentation results. Further post-processing algorithms to convert raw classification results on the snippet level to segmentation results can be developed, but we did not commence it. Instead, we developed a straight-forward learning from observation algorithm facilitate noisy snippet-level classification result.

% Overview of algorithm (IRL-(BC)-RL)
The proposed learning from observation procedure consists of three consecutive components: (1) reward function inference, utilizing noisy snippet level labels for each step, (2) behavioral cloning (BC), that exploits state-action pairs from the demonstration (if available) to expedite policy learning, and (3) reinforcement learning, working on top of the inferred reward function and the initial policy produced by 1 and 2, respectively. 
%The BC step is done via supervised training with state-action pairs, and proximal policy optimization (PPO) algorithm is used for RL.

% BC-and-RL is not our target.
%Since the BC and RL parts do not utilize the auxiliary videos clips we gathered through the activity localization algorithm, these two parts are not specifically our target of interest in this paper. Therefore, we adopted the most close-at-hand approaches for these two steps; the BC step is done by supervised training with state-action pairs, and proximal policy optimization (PPO) algorithm is used for RL.

%Now, we explain the inverse reinforcement learning algorithm that can infer reward function utilizing videos with noisy snippet level predictions. We can naturally expect that accumulated rewards should increase in a successful execution (or in a successful demonstration) since we expect incremental progress on a task.
%Therefore, if we train a neural network $g$ that can predict progress (outputs a single value representing a task completion rate), we can make use of it as a surrogate reward function.
In order to perform reward function inference with the auxiliary snippets for each activity, we utilize a shuffle-and-learn-style loss \cite{shuffle-misra2016unsupervised}. Intuitively, we expect that accumulated rewards should (approximately) monotonically increase in a successful execution (or in a successful demonstration) of an activity. Based on this assumption, we can train a neural network $g$ that can predict progress (i.e. outputs a single value representing the rate of activity completion), using it as a surrogate reward function.

By asking the network to predict the order of two frames from a video, we can directly model the monotonic progress of an activity. This loss can be formally written as:
\begin{equation}
    Loss=L_{ce}(sigmoid(g(o_{t},o_{t'})),\mathds{1}(t<t')),
\end{equation}
where $g$ is a neural network, $o_t$ and  $o_{t'}$ are observations (video frames) of the same activity at each time step $t$ and  $t'$, and $L_{ce}$ is the cross entropy loss.
This formulation is similar to both Shuffle-and-Learn \cite{shuffle-misra2016unsupervised} and TCN \cite{TCN2017-sermanet}; however, these methods use a triplet-based loss, whereas we have found empirically that using simple ordered pairs yields improved performance in our setting. Additionally, our approach does not require the additional hyperparameter that TCN uses to construct the positive and negative example for each triplet.

%while our work uses a pair of observations, the original versions use triplets. Since a triplet is a tuple of an anchor data point and corresponding positive and negative examples, an equivalence metric over data points is required to construct the triplet. While Sermanet et al. \cite{TCN2017-sermanet} circumvent this problem by adopting a new hyperparameter defining equivalence in the time domain, by using a pair of data samples, we only require a notion of order, which is well defined in the time domain.  \SN{FIX THIS AFTER TALKING TO WONJOON}

By measuring the progress of an activity through the trained function $g$, we can provide a surrogate reward signal to the agent; it gets a positive reward if it makes forward progress and vice versa. Many formulations of the function $g$ are possible, such as $sigmoid(g(o_{t-1},o_{t}))-0.5$, directly measuring progress between current time step and the previous time step. Yet, we empirically found that measuring progress between very adjacent frames is unstable. Instead, we measure progress with an initial frame $o_0$ as an anchor, then use a difference of raw $g$ values as a reward function:
\begin{equation}
    R_t = g(o_0,o_{t+1}) - g(o_0,o_{t}).
\end{equation}

% Advantage of our approach: no fancy segmentation algorithm is required!
%The training of the function $g$ does not require a segmented video; it only requires a pair of video frames extracted from a same video that belong to the same step of the task. Therefore, we can use the raw snippet-level label without further post processing to get a segmentation for each step. Furthermore, we empirically observed that trained $g$ function is robust to noise derived from noisy prediction the activity localization algorithm made.

We train the function $g$ for each activity by sampling pairs of video frames that both come from the same video and have the same predicted activity label. We empirically observed that trained function is fairly robust to noisy labels from the activity localization algorithm, as we will further discuss in the experiments. Furthermore, unlike the work of Sermanet et al. \cite{TCN2017-sermanet}, the proposed reward function does not require time-aligned video demonstrations as input. This simple but effective algorithm for reward function inference is an additional contribution.

\section{Evaluation}
% Two experiments are done: activity localization & IRL-RL pipeline
We conducted two types of experiments to assess the performance of our algorithms.
First, an action localization experiment measured the performance of the the proposed algorithms in both well-constrained simulated videos and unconstrained real videos.
Second, a policy learning experiment demonstrated the importance of the activity localization step for a multi-step task.

\subsection{Activity Localization Experiment: Setup}

% Introduce Simulated Reacher / Real-World ActivityNet
We examined the proposed meta-learning based activity localization approach with both simulated and real-world videos.
The simulated videos featured a two-joint robotic arm performing a reaching task. These videos were relatively consistent in content and presentation, making them, in principle, easier to analyze than unconstrained videos. 
By contrast, real-world videos present more challenges, including highly variable camera angles and environmental features. In our experiment, ActivityNet \cite{caba2015activitynet}, which is commonly used for activity classification or detection, is adopted.

\begin{figure*}[!tb]
    \centering
    \begin{subfigure}[t]{\linewidth}
        \centering
        \includegraphics[width=0.9\textwidth,height=0.08\textheight]{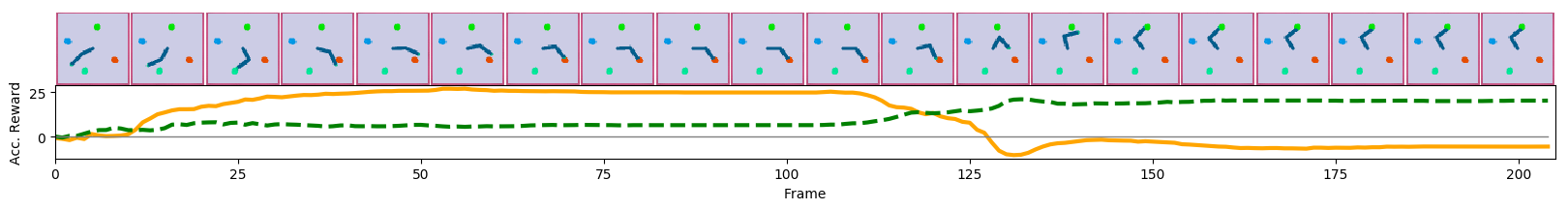}
        \caption{
            Accumulated reward under the reward functions inferred by our algorithm for the orange subtask (orange line) and the green subtask (green dashed line) during an execution of the task in which the agent reaches first for the orange target and then for the green target. %The reward function is inferred with action localization results predicted by MAML method. Accumulated rewards go up only when robot's end-effector reaches to a correct target.
        }
        \label{fig:IRL_results_1}
    \end{subfigure}
    \\
    \begin{subfigure}[t]{\linewidth}
        \centering
        \includegraphics[width=0.9\textwidth,height=0.1\textheight]{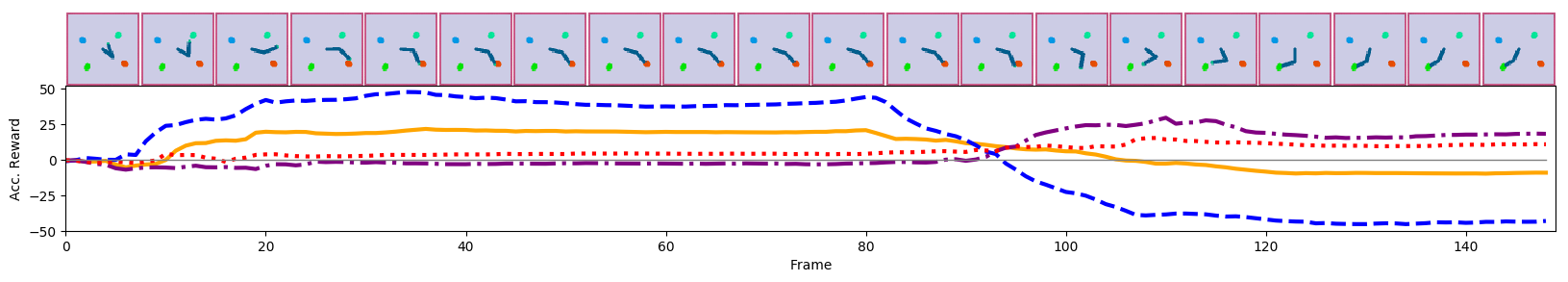}
        \caption{
            The agent reaches for the orange target, followed by the green target. The plot shows accumulated rewards under the reward function for the orange subtask when learning was performed with ground-truth video segmentations (blue dashed line), MAML segmentations (orange solid line), a single demonstration video (purple dash-dot line), and whole videos without any segmentation (dotted red line). While the reward function from MAML-segmented videos displays a similar pattern as that of ground-truth (increasing during movement toward orange,  decreasing during movement toward green), the reward function inferred from only a single demonstration video and unsegmented videos does not, since those functions are either inferred from too little data or with both relevant and irrelevant frames.
        }
        \label{fig:IRL_results_2}
    \end{subfigure}
    \caption{Reward function inference results with validation set videos on two-joint reacher environment.}
    \label{fig:IRL_results}
    \vspace{-0.5cm}
\end{figure*}

\textbf{Reacher environment}
This dataset contains simulated videos of a two-joint robot arm trying to select and reach for targets based on color. In total, 4 potential targets of different colors are present, but the arm must reach for the correct 2 or 3 target colors, depending on the number of steps in the task.
These videos were created using the Bullet physics simulator \cite{bullet} with a pretrained policy that can successfully reach a target position.
Note that the task involves selecting a set of multiple target colors in any order, so different videos featuring the robot solving the same underlying task (such as \{orange, green\}) can present the subtasks in a different order. Some sample videos are shown in Fig.~\ref{fig:IRL_results}.

We generated three datasets of videos with different configurations.
For the first, most basic dataset, we generated videos with 4 colors, using 2 of them as targets (6 total combinations). For each combination, we generated 100 videos. Also, meta-test set videos (of unseen activities) were generated with 4 colors different from those in the training and validation sets.
The second, more complex dataset was generated from a set of 6 possible colors, having a task of length 3.
However, only 4 colors are shown in each scene---three target colors and one distractor color. Meta-test videos were also generated in the same manner, but with a set of 6 different colors. For each possible color combination, we generated 40 videos.
The thrid dataset was designed to test learning transfer in significantly novel settings. Length 2 tasks with 36 colors were generated and used as a training and a validation set, with 3 videos for each possible task. Then the meta-trained network was tested on: videos with (1) another set of 36 different colors, (2) a bad reacher policy (under-damping around a target) with 4 different colors, and (3) a three-joint reacher arm with 4 different colors. 
The resolution of the videos was 64 by 64 pixels, and we used snippets of 16 raw video frames as an input for each network. 

\textbf{ActivityNet}
This dataset includes video clips of 100 human activities, such as playing golf or drinking coffee. While our algorithm is targeted to a single task video having multiple activities, the videos in this dataset only contain a single activity type. However, we can perform the same activity localization task by creating \textit{pseudo} tasks by concatenating single-activity videos. This was done by randomly choosing 5 activities and using a random video segment from each activity. %A random label was assigned to each activity for every draw task.
%so the meta-learner have to deduce the way to fast adopt to a given video segments and labels so that it can perform correct prediction respecting given labels. 
%It is slightly contrived, but it was adopted as a proof-of-concept, demonstrating the algorithm’s capability to work on real world videos.
This would not be a good baseline for a segmentation algorithms, since there are stark differences between activity changes; however, given that our approach uses snippet-based activity classification, this does not give our algorithm an unfair advantage, and thus serves as a proof-of-concept that our approach can work on real-world datasets.  However, this does suggest that our approach could be improved in future work by integrating our snippet-based method with more traditional segmentation that takes the broader temporal context into account.

We chose 80 activities for meta-training, and the remaining 20 activities were held out for meta-testing. We chose 5 random activities to make a pseudo multi-step task video, such that 5-way classification tasks were generated.
%Even though we could treat each instance of an activity independently and let a neural network generate a single prediction per a video, a snippet-level dense prediction was made in the consideration of real setting.
3D ConvNet \cite{tran2015learning} feature vectors of length 500 were extracted for every 16 frames for each video, and each of the feature vectors were used as an input snippet for our algorithm.
Because the same set of activities were displayed across demonstration and target videos in both datasets, we set the prediction suppressing threshold low-enough so that no ``none of the above'' predictions we made.

\textbf{Network architectures and baselines}
For the reacher environment, a three-layer 3D CNN was used across different configurations as a base architecture, and only the final classification layer was modified, depending on the number of steps in a target task. For example, for the base reacher environment with two steps, a fully connected layer outputting two logits was added on top of the base architecture. We used our MAML-based framework for training. Detailed hyperparameters used in the experiment can be found in the publicly available code\footnote{\href{https://github.com/hiwonjoon/ICRA2019-Activity-Localize}{https://github.com/hiwonjoon/ICRA2019-Activity-Localize}}.

A neural network sharing the same base network architecture was used as a baseline. It was trained with classification objective but with all the possible labels existing in the training set videos: the number of colors in its configuration. Then, we removed the last classification layer and used the trained network as a feature extractor. Activity localization was performed by assigning a label of the closest demo snippet to a target snippet in the feature space via Euclidean distance.

For the ActivityNet dataset, we used a recurrent neural network (RNN) with gated recurrent units (GRU) \cite{cho-gru}. Specifically, two fully connected layers were applied to snippet-level feature vectors, and these outputs was passed through the RNN, followed by a final classification layer to make a dense prediction. In this setup, we used our Reptile-based framework for computational reasons. Since Reptile does not require second order derivatives, not only can the training be performed quickly, but also it does not blow up the size of computation graph when unrolling over many time steps.

A simple RNN classifier was adopted as a baseline. The RNN was trained with a classification loss sharing the same base architecture, and it was used as a feature extractor in the same fashion as the reacher environment. The activity localization was also performed in the same way as in the reacher experiments.

\subsection{Activity Localization Experiment: Results}

\begin{table}[tb]
   % \begin{subtable}{0.5\linewidth}
      \centering
      \captionsetup{width=.9\linewidth}
      \begin{tabular}{|c|cc|}
        \multicolumn{3}{c}{Panel A: Reacher \footnotesize{(2 subtasks, 4 colors each)}}\\
        \hline
                    & Classifier & Meta Learning      \\ \hline
        Same Task   & 0.4820 & \textbf{0.8479} \\
        Unseen Task & 0.4514 & \textbf{0.6944} \\ \hline
        \multicolumn{3}{c}{}\\
        \multicolumn{3}{c}{Panel B: Reacher \footnotesize{(3 subtasks, 6 colors)}}\\
        \hline
                    & Classifier & Meta Learning      \\ \hline
        Same Task   & 0.7410 & \textbf{0.7852} \\
        Unseen Task & 0.6512 & \textbf{0.7189} \\ \hline
        \multicolumn{3}{c}{}\\
      \end{tabular}
      \begin{tabular}{|c|cc|}
        \multicolumn{3}{c}{Panel C: Reacher \footnotesize{(2 subtasks, more colors)}}\\
        \hline
                    & Classifier    & Meta Learning      \\ \hline
        Same Task \footnotesize{(36 colors)} & 0.7418 &  \textbf{0.7867} \\
        Unseen Task \footnotesize{(36 colors)} & 0.6749 & \textbf{0.7015} \\ \hline
        Bad Policy Arm & 0.5432 & \textbf{0.6337} \\
        Three-joint Arm & 0.5218 & \textbf{0.6161} \\ \hline
      \end{tabular}
      \begin{tabular}{|cc|cc|}
        \multicolumn{4}{c}{}\\
        \multicolumn{4}{c}{Panel D: ActivityNet Dataset}\\
        \hline
         &  & \begin{tabular}[c]{@{}c@{}}RNN\\ Classifier\end{tabular} & \begin{tabular}[c]{@{}c@{}}RNN\\ Reptile\end{tabular} \\ \hline
        \multicolumn{1}{|c|}{\multirow{2}{*}{mIoU}} & Validation set & 0.2121 & 0.3585 \\
        \multicolumn{1}{|c|}{} & Meta-test Set & 0.2245 & 0.2883 \\ \hline
        \multicolumn{1}{|c|}{\multirow{2}{*}{Accuracy}} & Validation Set & 0.2266 & 0.4894 \\
        \multicolumn{1}{|c|}{} & Meta-test Set & 0.2428 & 0.4077 \\ \hline
      \end{tabular}
%      \begin{tabular}{|cc|cc|}
%        \multicolumn{4}{c}{}\\
%        \multicolumn{4}{c}{Panel E: Breakfast Dataset}\\
%        \hline
%        &  & \begin{tabular}[c]{@{}c@{}}RNN\\ Classifier\end{tabular} & \begin{tabular}[c]{@{}c@{}}RNN\\ Reptile\end{tabular} \\ \hline
%        \multicolumn{1}{|c|}{\multirow{2}{*}{mIoU}} & \begin{tabular}[c]{@{}c@{}}Validation\\ Set\end{tabular} & 0.1841 & 0.2220 \\
%        \multicolumn{1}{|c|}{} & \begin{tabular}[c]{@{}c@{}}Meta-test\\ Set\end{tabular} & 0.1854 & 0.2223 \\ \hline
%        \multicolumn{1}{|c|}{\multirow{2}{*}{Accuracy}} & \begin{tabular}[c]{@{}c@{}}Validation\\ Set\end{tabular} & 0.1462 & 0.3103 \\
%        \multicolumn{1}{|c|}{} & \begin{tabular}[c]{@{}c@{}}Meta-test\\ Set\end{tabular} & 0.1512 & 0.3184 \\ \hline
%      \end{tabular}
    \caption{Activity localization results on Reacher-domain and the real-world ActivityNet dataset; mIoU is displayed.}
    \label{table:alignment-results}
    \vspace{-0.5cm}
\end{table}

% Brief interpretation of the data
Our results are presented in Table~\ref{table:alignment-results}.
Our meta-learning methods showed significantly better results than the baseline classifier for both the simulated and real world domains, even when using the validation sets for which the baseline was directly optimized. Though a slight drop in the performance can be observed for all meta-test set cases, it still demonstrates that the suggested method is able to adapt quickly without requiring a large video dataset for a novel task. 
Though the mean intersection-over-union (mIoU) performance on real videos dropped noticeably compared to the simpler simulated videos, there is large room for improvement since we employed a very simple network for this proof-of-concept experiment. Future work may also include the exploration of methods to suppress the high frequency noise in dense predictions, as well as techniques to more directly handle the large variance in real-world video via specialized network architectures like TCN \cite{TCN2017-sermanet}.

% \begin{figure}[!htbp]
%     \centering
%     \includegraphics[width=0.48\textwidth]{assets/Breakfast-Detailed.pdf}
%     \caption{
%         mIoUs per task for three different methods.
%     }
%     \label{fig:breakfasst-detailed}
% \end{figure}

% \begin{figure*}[!htbp]
%     \centering
%     \includegraphics[width=\textwidth]{assets/Aligned_Results_Breakfast/cereal_8_0.png}
%     \caption{
%         A demo video and a sample video from the breakfast dataset are shown, along with ground-truth segmentation.
%         An segmentation of the sample video to the demo video using the metalearning method, triplet network, and classifier baseline follows.
%         Colors correspond to: brown (take bowl), blue (pour cereal), green (pour milk), grey (stir cereal).
%         %Best viewed in electronic copy.
%     }
%     \label{fig:breakfasst-aligned-example}
% \end{figure*}

\subsection{Policy Learning with Video Snippets: Setup}

Next, we evaluated the the full multi-step learning pipeline in the two-joint reacher arm environment discussed in the previous subsection. This experiment is designed to show (1) the importance of activity localization for multi-step task learning and (2) the robustness of our reward function inference approach to noise in the localization step.

We generated 799 new videos for the \{orange, green\} task, and snippets of the videos were classified using the meta-trained model from the previous section, with one demonstration video as input.
To investigate how the noise in the predicted snippet labels might affect reward function inference (compared to using perfectly segmented videos) and policy learning (compared to having ground truth rewards), we compared the performance of policies based on reward functions inferred from MAML with policies learned from (1) ground truth rewards (upper bound for RL performance), (2) rewards inferred from perfectly segmented videos (upper bound for the suggested reward function learning method), and (3) rewards inferred from unsegmented videos (baseline). Reward function inference was performed via the shuffle-and-learn-style LfO approach described in Section \ref{sec:lfo}.

Reward function inference was conducted by training the proxy function $g$. We designed a two-stage neural network for $g$: a frame embedder and a progress predictor. The frame embedder converts a raw RGB image into a fixed-length embedding vector, and the progress predictor takes two such embeddings as input and emits a single logit value representing the relative order between the two input frames. A neural network having three 2D convolutional layers followed by two fully connected layers was used for the frame embedder, and a two-layer, fully connected neural network was used for the progress predictor. The embedding feature vector length was 64. Both of the networks were trained in a end-to-end fashion with the aforementioned loss functions. Detailed hyperparameters, such as kernel size, can be found in the publicly available code.

The policies were trained using proximal policy optimization (PPO) \cite{ppo-schulman2017} using the implementation from OpenAI baselines \cite{baselines-openai}.
A two-layer, fully connected neural network was used to represent the policy, with identical hyperparameters for each of the policies.
The performance of a learned policy was measured by success rate among 300 trials; if the robot reached a target location and remained there for more than 32 frames, then a policy rollout was regarded as a success.

\subsection{Policy Learning with Video Snippets: Results}

\textbf{Reward Function Inference Results}
In Figure~\ref{fig:IRL_results}, we show the accumulated reward from the  inferred reward function of each subtask during a successful task execution (\ref{fig:IRL_results_1}), as well as a comparison of accumulated reward for the orange subtask when the reward function is inferred using different segmentation methods (or no segmentation at all) (\ref{fig:IRL_results_2}).
First, we can clearly observe that each of the trained models effectively represents the degree of completion for each different subtask; the value goes up only as it reaches the correct target.
The results also confirmed the necessity of video segmentation---without segmentation, the inferred reward function did not correctly represent any of the subtasks individually, nor the overall task. 

%A noisy reward function cannot provide a meaningful policy.
%In contrast, noise-free videos generated the most stable reward functions.

\begin{figure}[tb]
    \centering
    \includegraphics[width=0.8\linewidth]{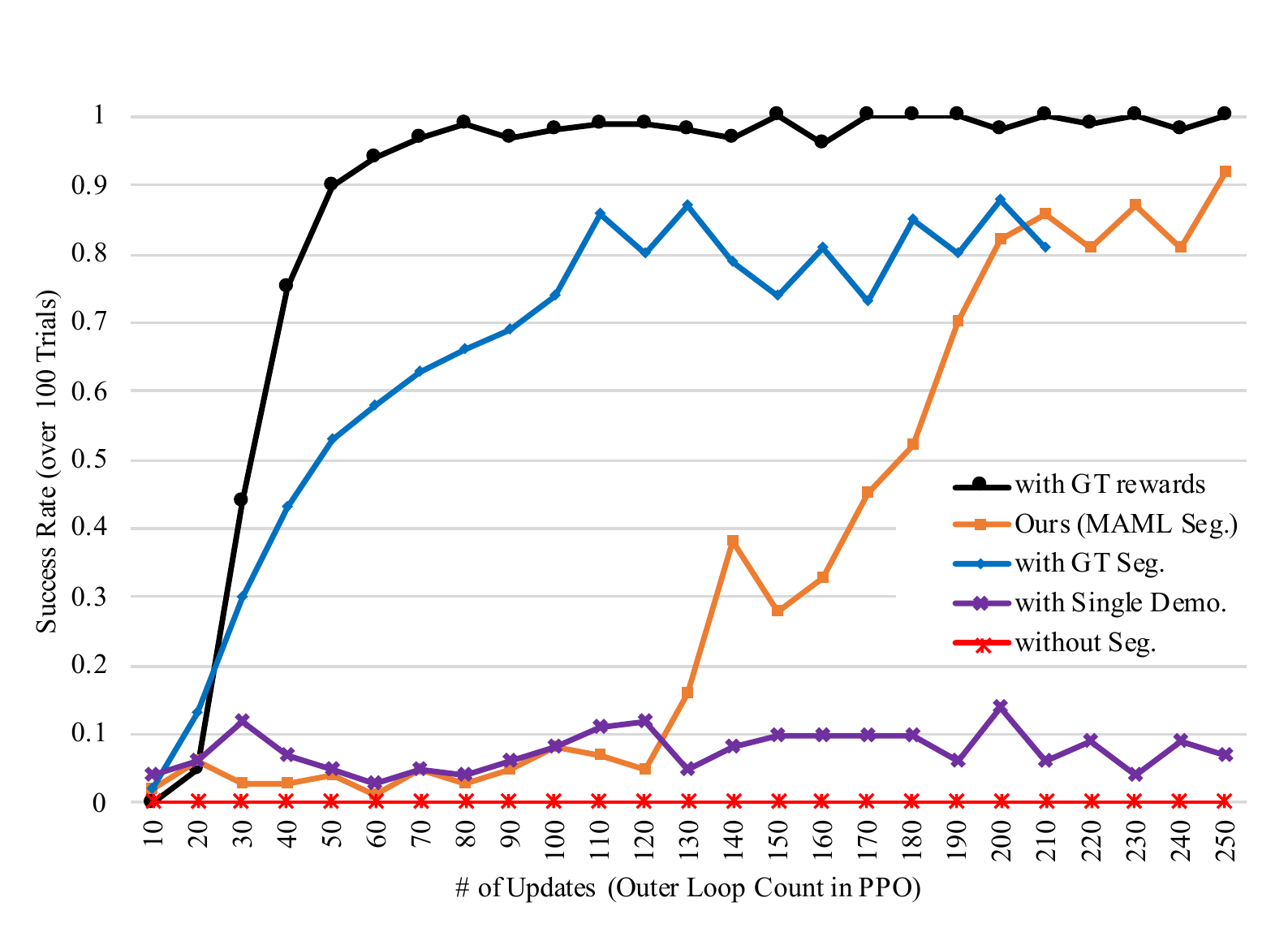}
    \caption{
        The policy success rate vs. number of PPO iterations.
        Reward functions from 5 different sources for the 2-joint reacher task (orange target) are shown.
    }
    \label{fig:ppo_learning_statistics}
    \vspace{-0.5cm}
\end{figure}

\textbf{RL Results}
The results of the reinforcement learning experiments are shown in Table~\ref{table:ppo-result} and Figure~\ref{fig:ppo_learning_statistics}.
We confirmed that inferred reward functions based on only raw video frames can be used to generate meaningful policies for multi-step tasks, but only when videos are properly divided into subtasks, as the reward signal learned without separation does not result in a successful policy. 
Furthermore, it was also possible to generate a successful policy with noisy activity predictions from the proposed meta-learning method, rather than requiring perfect segmentation of a video. 
Interestingly, even with the imperfect activity localization that our method generated, it (asymptotically) resulted in a policy that performed as well as the perfectly segmented case, providing evidence that this approach may work with noisy localization on real-world video as well.
Furthermore, it can be seen that performance is poor when only learning from a single demonstration; without auxiliary data, there is not enough data to enable generalization, motivating our approach.
%It is also noticeable that the inferred reward function with a single demonstration can result in a policy that solves the small subset of the problem. This verifies our motivation on using auxiliary video data for achieving generalization.

% SCOTT: I think this hurts our case more than it helps it, so I took it out
%However, Figure~\ref{fig:ppo_learning_statistics} also illustrates the advantage of high-quality segmentation. 
%The policy improved more quickly with the inferred reward function based on ground-truth segmentation, compared to one inferred from our activity localization approach.
%Additionally, it was qualitatively noticeable that the policy trained with the reward function based on perfectly segmented videos generated more stable, smooth actions than the policy trained based on our method.
%Finally, the policy using perfectly segmented videos has a stronger qualitative resemblance to a policy trained using the ground truth reward function.
%These results reinforce the motivation proposed by this paper: accurate separation of subtasks is essential for learning from demonstration of a multi-step task.

\begin{table}[htb]
    \centering
    \captionsetup{width=\linewidth}
    \renewcommand{\arraystretch}{1.3}
    \begin{tabular}{|c|cc|}
        \hline
                          & Target Orange    & Target Green     \\ \hline
        with GT rewards   & 0.9600 \footnotesize{(288/300)} & 0.9833 \footnotesize{(295/300)} \\
        Ours (MAML)       & 0.8533 \footnotesize{(256/300)} & 0.6800 \footnotesize{(204/300)} \\
        with GT Seg.      & 0.7867 \footnotesize{(236/300)} & 0.8300 \footnotesize{(249/300)} \\
        with Single Demo. & 0.04 \footnotesize{(12/300)} & 0.07 \footnotesize{(21/300)} \\
        without Seg.      & 0.0 \footnotesize{(0/300)} & 0.0 \footnotesize{(0/300)} \\ \hline
    \end{tabular}
    \caption{
        Reinforcement learning results with inferred reward functions and PPO algorithm.}
    \label{table:ppo-result}
    \vspace{-0.5cm}
\end{table}

\section{Conclusion}
\label{sec:conclusion}

We addressed the challenging problem of learning multi-step tasks in a one-shot LfO setting by introducing novel algorithms that (1) perform one-shot activity localization in auxiliary videos to provide additional examples of each activity, and (2) infer reward functions for each step of the task individually, using the action-localized videos. We first examined the activity localization algorithm on both simulated and real-world datasets, showing that our approach can successfully classify snippets of a video into activities defined by a single segmented demonstration video. 
%Second, as the first step toward more general learning from observation algorithms, we used a new LfO pipeline that leveraged activity-specific video snippets from our activity localization algorithm. 
Second, our full proposed learning pipeline was tested on a multi-step reaching task. Our proposed method successfully completed these tasks, while the baseline LfO methods---which did not take the multi-step nature of the task into account or did not use auxiliary data---were shown to fail.

We anticipate that this work will serve as a step toward more general LfO algorithms that will be able to fully leverage unstructured web-scale video data for complex, multi-step tasks in the future. 
To achieve this, the most important piece of future work is scaling up this framework to work with real-world video data.  This will present several challenging issues such as, including coping with visual differences between environments that auxiliary videos are captured in and handling multiple viewpoints or unsteady egocentric video.
Another promising direction of future work is exploiting additional data modalities, such as text-based descriptions of tasks or audio signals from videos that can help to detect, segment, and infer reward functions from actions in a video. 

%Additionally, there are many opportunities to improve aspects of LfO algorithms in the future.  
%Additionally, while we were able to obtain excellent sample efficiency in terms of number of demonstrations required to infer reward functions, we did not consider the sample complexity of the RL step that followed.  %Future LfO algorithms may, for example, be able to reduce this burden by regressing from a single demonstration (and auxiliary video data) directly to a policy, skipping the RL step entirely.
%Future work may focus

%The proposed pipeline was able to achieve sample efficiency for a ``demonstration'', but the sample efficiency of ``reinforcement learning'' is not considered. Directly deriving a policy from a raw observation without RL algorithm or learning environment models instead of inferring reward signal could be a good solution to this problem.

%%%%%%%%%%%%%%%%%%%%%%%%%%%%%%%%%%%%%%%%%%%%%%%%%%%%%%%%%%%%%%%%%%%%%%%%%%%%%%%%



\section*{ACKNOWLEDGMENT}

%The preferred spelling of the word �acknowledgment� in America is without an �e� after the �g�. Avoid the stilted expression, �One of us (R. B. G.) thanks . . .�  Instead, try �R. B. G. thanks�. Put sponsor acknowledgments in the unnumbered footnote on the first page.
This  work  has  taken  place  in  the  Personal  Autonomous Robotics Lab (PeARL) at The University of Texas at Austin. PeARL  research  is  supported  in  part  by  the  NSF  (IIS-
1724157, IIS-1638107, IIS-1617639, IIS-1749204) and ONR (N00014-18-2243).

\clearpage

%%%%%%%%%%%%%%%%%%%%%%%%%%%%%%%%%%%%%%%%%%%%%%%%%%%%%%%%%%%%%%%%%%%%%%%%%%%%%%%%

\bibliographystyle{IEEEtran}
\bibliography{references}

\end{document}